\documentclass{article}

\usepackage{arxiv}
\usepackage[utf8]{inputenc}  
\usepackage[T1]{fontenc}     
\usepackage{hyperref}        
\usepackage{url}             
\hypersetup{breaklinks=true}
\usepackage{ragged2e}
\usepackage{booktabs}        
\usepackage{amsfonts}        
\usepackage{nicefrac}        
\usepackage{microtype}       
\usepackage{graphicx}
\graphicspath{ {./Figures/} }
\usepackage{float}
\usepackage{subfig}
\usepackage{enumitem}
\usepackage{longtable}

\title{ A brief history of AI: how to prevent another winter (a critical review)}

\author{
 Amirhosein Toosi \\
  Department of Integrative Oncology\\
  BC Cancer Research Institute\\
  Vancouver, BC, V5Z 1L3 \\
  \texttt{atoosi@bccrc.ca} \\
  \And
 Andrea Bottino \\
  Department of Computer and Control Eng.\\
  Polytechnic University of Turin\\
  Turin, Italy 10129 \\
  \texttt{andrea.bottino@polito.it} \\
  \And
 Babak Saboury \\
  Department of Radiology and Imaging Sciences\\
  National Institutes of Health\\
  Bethesda, MD 20892 \\
  \texttt{babak.saboury@nih.gov} \\\
  \And
 Eliot Siegel \\
  Department of Diagnostic Radiology and Nuclear Medicine\\
  University of Maryland School of Medicine\\
  Baltimore, MD 21201 \\
  \texttt{esiegel@umaryland.edu} \\
  \And
 Arman Rahmim \\
  Departments of Radiology and Physics\\
  University of British Columbia\\
  Vancouver, BC \\
  \texttt{arman.rahmim@ubc.ca} \\
}
\begin{document}
\textbf{\textit{This manuscript has been accepted for publication in PET Clinics, Volume 16, Issue 4, P449-469, Oct. 01, 2021}}
DOI:\url{https://doi.org/10.1016/j.cpet.2021.07.001}
\vspace{400pt}
\maketitle
\begin{abstract}
The field of artificial intelligence (AI), regarded as one of the most enigmatic areas of science, has witnessed exponential growth in the past decade including a remarkably wide array of applications, having already impacted our everyday lives. Advances in computing power and the design of sophisticated AI algorithms have enabled computers to outperform humans in a variety of tasks, especially in the areas of computer vision and speech recognition. Yet, AI’s path has never been smooth, having essentially fallen apart twice in its lifetime (‘winters’ of AI), both after periods of popular success (‘summers’ of AI). We provide a brief rundown of AI's evolution over the course of decades, highlighting its crucial moments and major turning points from inception to the present. In doing so, we attempt to learn, anticipate the future, and discuss what steps may be taken to prevent another ‘winter’.
\end{abstract}


\section{Introduction}

\begin{figure}[h]
	\centering    
    \includegraphics[clip,width=1\linewidth]{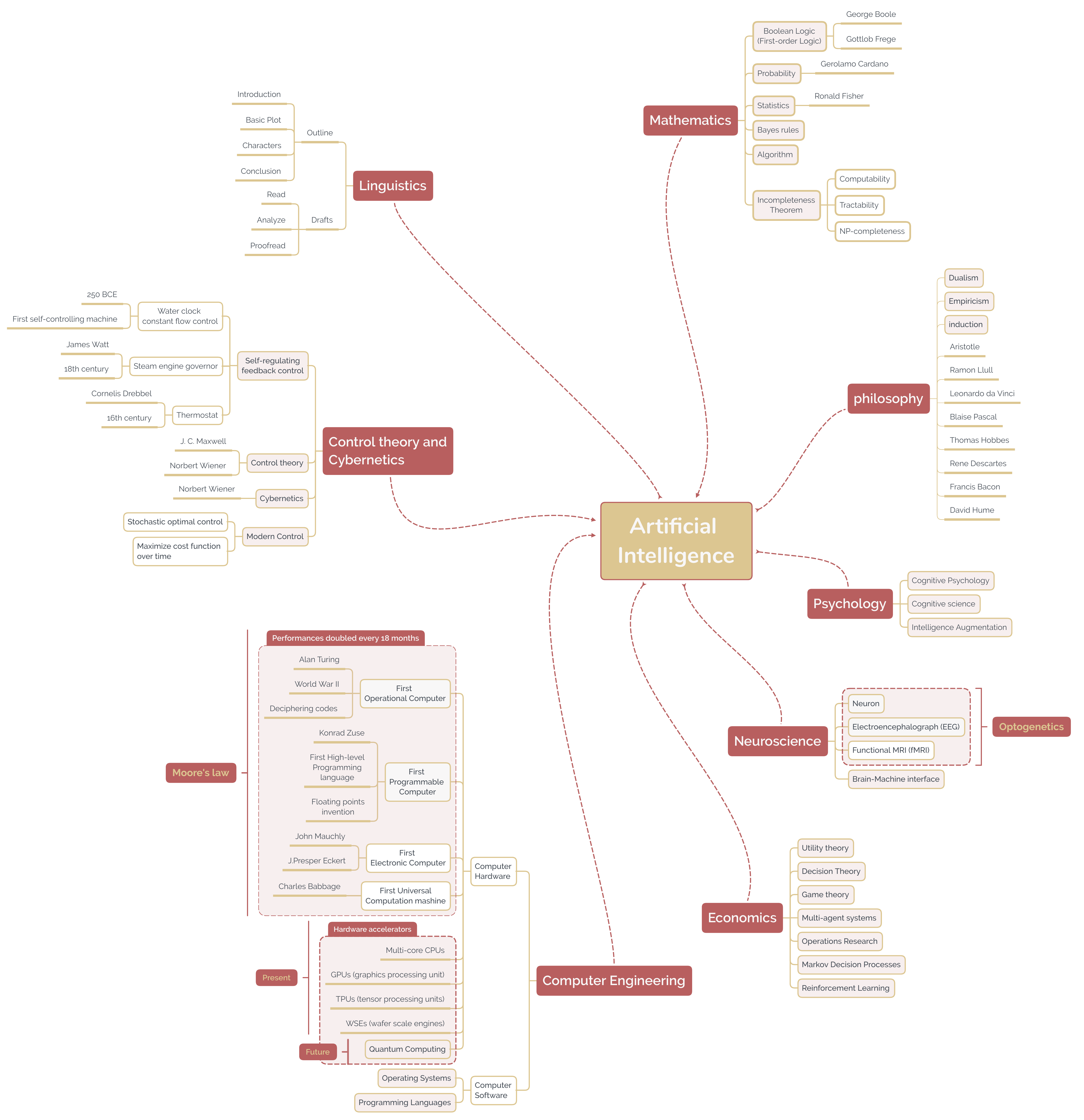}
	\caption{A summary of the influence of different fields on AI.}
	\label{fig:Fig1}
\end{figure}

Artificial Intelligence (AI) technology is sweeping the globe, leading to bold statements by notable figures: “[AI] is
going to change the world more than anything in the history of mankind” \cite{Catherine_Clifford2019-xz}, “it is more profound than even 
electricity or fire" \cite{Catherine_Clifford2018-bu}, and "just as electricity transformed almost everything 100 years ago, today I actually have
a hard time thinking of an industry that I don’t think AI will transform in the next several years'' \cite{Lynch2017-jm}. Every few
weeks there is news about AI breakthroughs. Deep-fake videos are becoming harder and harder to tell apart from real
ones \cite{BBC_News2020-xb} \cite{Vincent2021-sp}. Intelligent algorithms are beating humans in a greater variety of games more easily. For the first time
in history, in arguably the most complex board game (named “Go”), DeepMind’s AlphaGo has beaten the world champion. 
AI has been around for decades, enduring “hot and cold” seasons, and like any other field in science, AI developments
indeed stand on the shoulders of giants (see figure~\ref{fig:Fig1}). With these in mind, this article aims to provide a picture of
what AI essentially is and the story behind this rapidly evolving and globally engaged technology.  

\section{What is AI?}

AI is “the theory and development of computer systems able to perform tasks normally requiring human intelligence,
such as visual perception, speech recognition, decision-making, and translation between languages” \cite{noauthor_2020-md}. Marvin Minsky,
American mathematician, computer scientist, and famous practitioner of AI defines AI as “the science of making
machines do things that would require intelligence if done by men” \cite{Dennis2021-ve}. John McCarthy who coined the term “artificial
intelligence” in 1956, described it as "the science and engineering of making intelligent machines". IBM suggests that
“Artificial intelligence enables computers and machines to mimic the perception, problem-solving, and decision-making
capabilities of the human mind” \cite{IBM_Cloud_Education_undated-ys}. McKinsey \& Company explains it as a “machine’s ability to mimic human cognitive
functions, including perception, reasoning, learning, and problem-solving” \cite{Chui_undated-xr}.

\begin{figure}[h]
	\centering    
    \includegraphics[clip,width=1\linewidth]{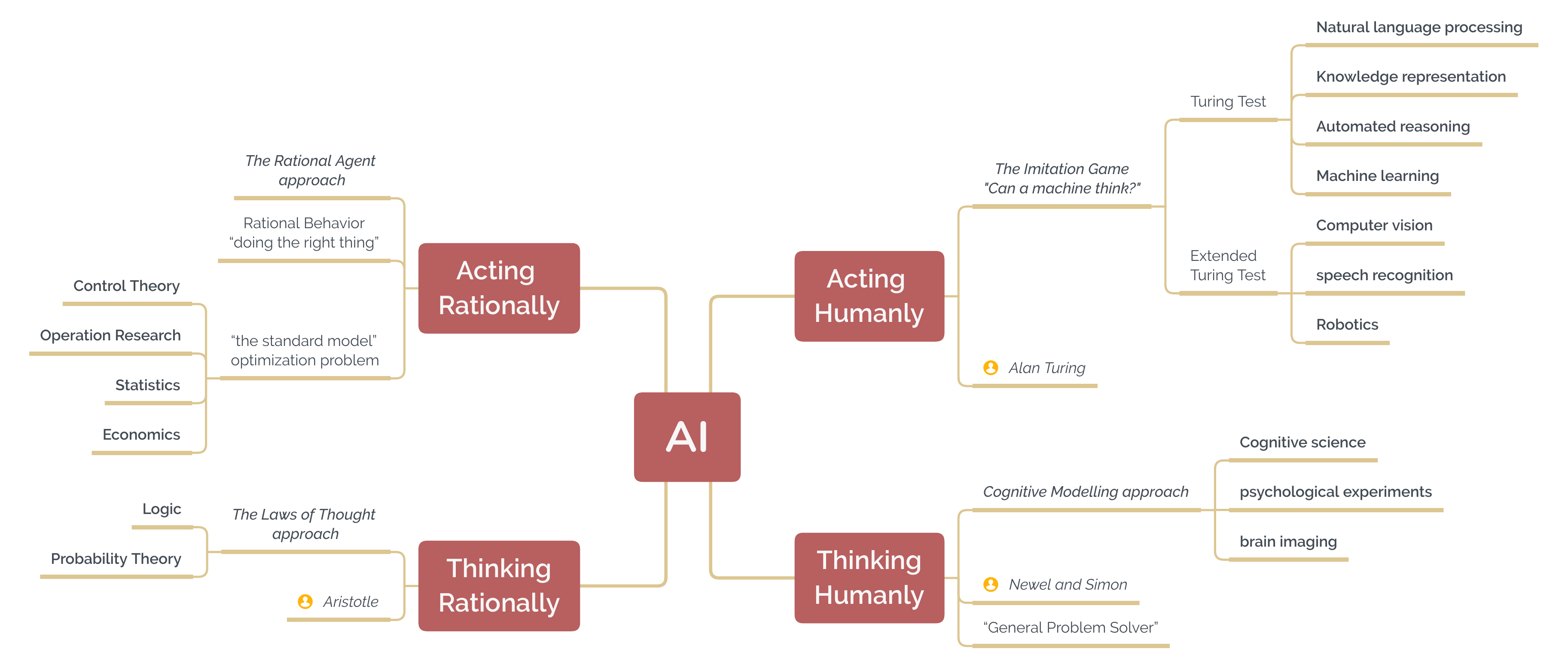}
    \caption{Summary of two-dimensional AI approaches as proposed by Russel and Norvig \cite{Russell2020-cx}
    (Data from Norvig, Peter, and Russell, Stuart. Artificial Intelligence: A Modern Approach, EBook, Global Edition. United Kingdom, Pearson Education, 2021.)}
    \label{fig:Fig2}
\end{figure}

Russel and Norvig \cite{Russell1995-yi} proposed four conceivable approaches to AI: acting humanly, thinking humanly, acting
rationally, and thinking rationally (see figure~\ref{fig:Fig2}). The British mathematician Alan Turing published a paper in 1950
(“Computers and intelligence” \cite{Turing1950-xt}) in which he proposed a tool to determine the difference between a task performed
by a person and a machine. This test, known as the "Turing test", consists of a series of questions to be answered. A
computer can pass the test if a human interrogator cannot tell whether the answers to the questions come from a person
or a computer. As such, to pass the test, the computer is required to have a number of essential capabilities such as: Natural language processing - to manage a natural and effective communication with human beings; Knowledge representation - to store the information it receives; Automated reasoning - to perform question answering and update the conclusion; And machine learning - to adjust to new situations and recognize new patterns.


In Turing’s view, a physical simulation of a human is totally irrelevant to demonstrate intelligence. Other
researchers, however, have suggested a complete Turing test \cite{Hernandez-Orallo2000-vo, Dowe1997-tr, Hayes1995-ju} that involves interaction with real-world objects
and people. Hence, the machine should be equipped with two additional (and vital) capabilities to pass the “Extended”
version of the Turing test \cite{Russell2020-cx}: Computer vision and speech recognition - to see and hear the environment; And robotics - to move around and interact with the environment. 

\section{History of AI}
\begin{figure}[h]
	\centering    
    \includegraphics[clip,width=1\linewidth]{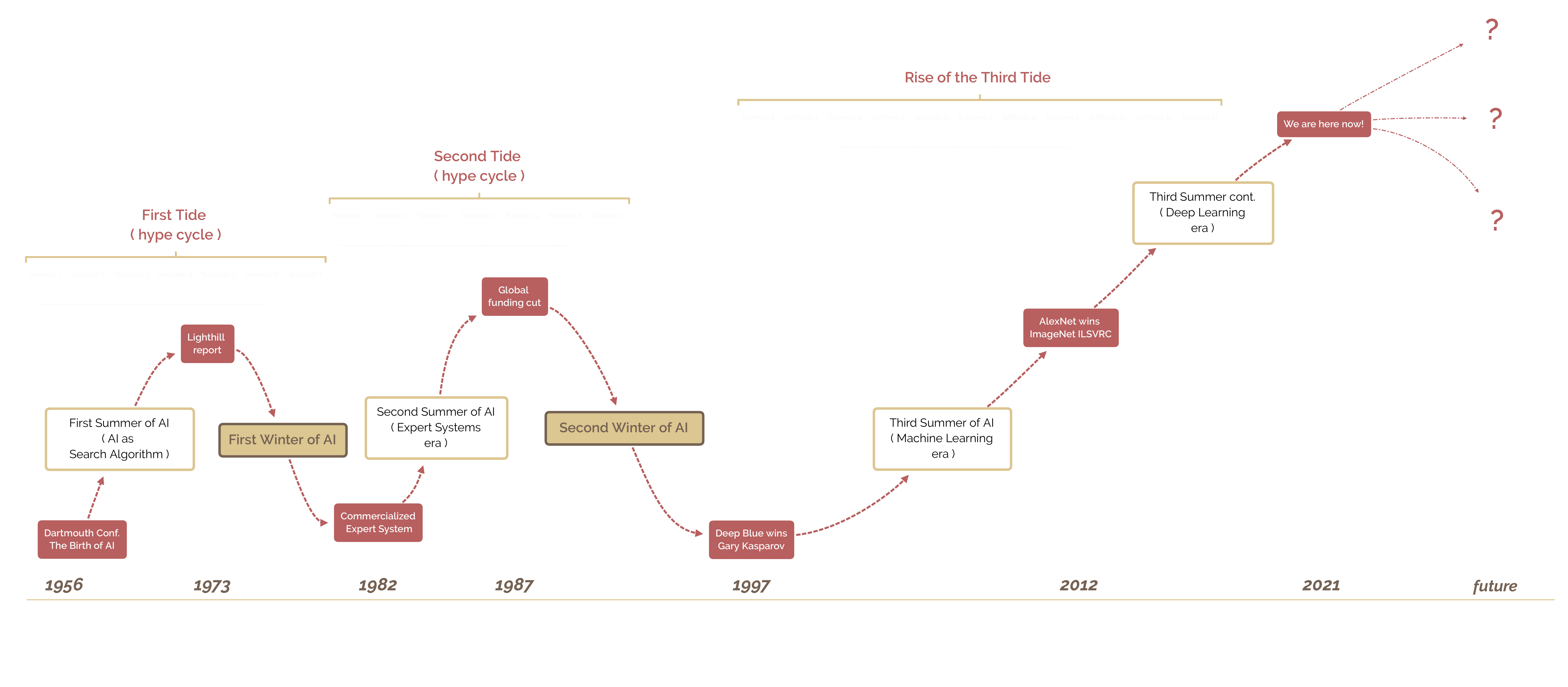}
\vspace{-2.5em}
	\caption{Dividing the history of AI into three recurring cycles (public excitements/time).(Adapted from T. Noguchi et al., “A practical use of expert system “AI-Q” focused on creating training data,” 2018 5th International Conference on Business and Industrial Research (ICBIR), 2018, pp. 73-76; With permission. (Figure 1 in original).)}
	\label{fig:Fig3}
\end{figure}

The field of AI has experienced extreme ascends and descends over the last seven decades. These recurring ridges of
great promise and valleys of disappointment referred to as AI's Summers and Winters, have divided the history of AI
into three distinct cycles (see figure~\ref{fig:Fig3}). These different cycles and seasons will be discussed in the following
parts.

\subsection{Prehistoric Events}
\begin{figure}[H]
    \centering
    \subfloat[\centering ]{{\includegraphics[width=6.11cm]{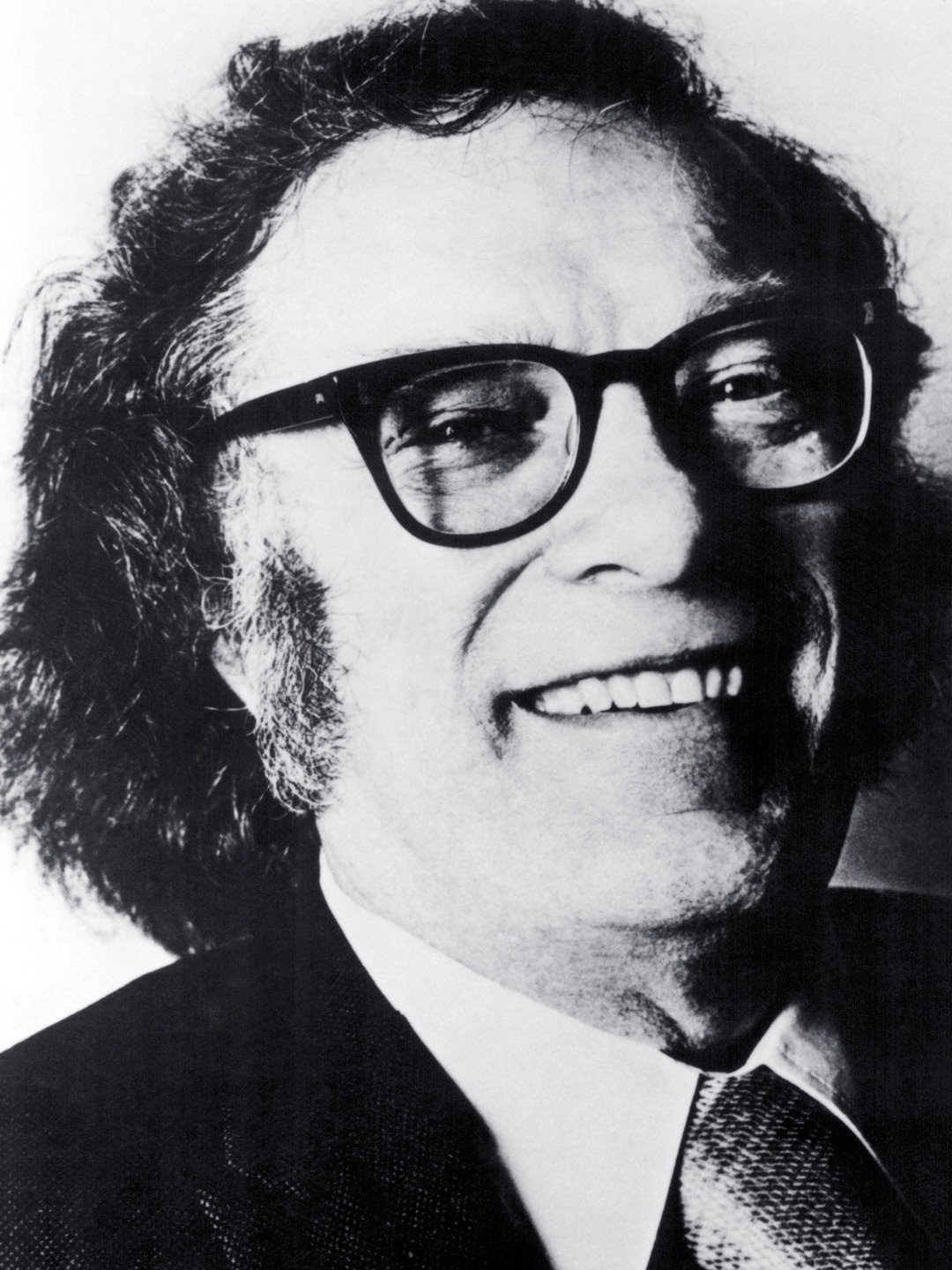} }}%
    \qquad
    \subfloat[\centering ]{{\includegraphics[width=5cm]{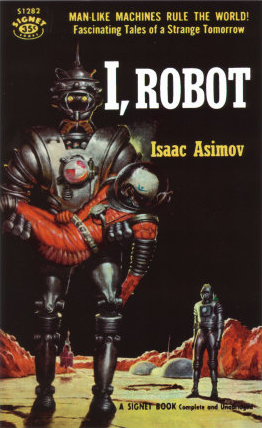} }}%
    \caption{(a): Isaac Asimov, the well-known sci-fi writer. (b): I, Robot, Asimov’s sci-fi
book series \cite{noauthor_undated-uv}.}%
    \label{fig:Fig4}%
\end{figure}

When science fiction writer Isaac Asimov wrote his timeless book "I, Robot" in 1942, he likely did not imagine that
this work, 80 years later, would become a primary source for defining the laws governing human-robot interactions
in modern AI ethics. Although Asimov's novels (figure~\ref{fig:Fig4}) are often considered as the birthplace of the ideas of
intelligent machines \cite{Haenlein2019-hu}, McCulloch and Pitts’ article, “A Logical Calculus of the Ideas Immanent in Nervous
Activity” published in 1943 \cite{McCulloch1943-xn}, was the first step toward the implementation of AI \cite{noauthor_undated-zt, noauthor_undated-oq, noauthor_undated-vc, Piccinini2004-cb}. 

\begin{figure}[h]
	\centering    
    \includegraphics[clip,width=1\linewidth]{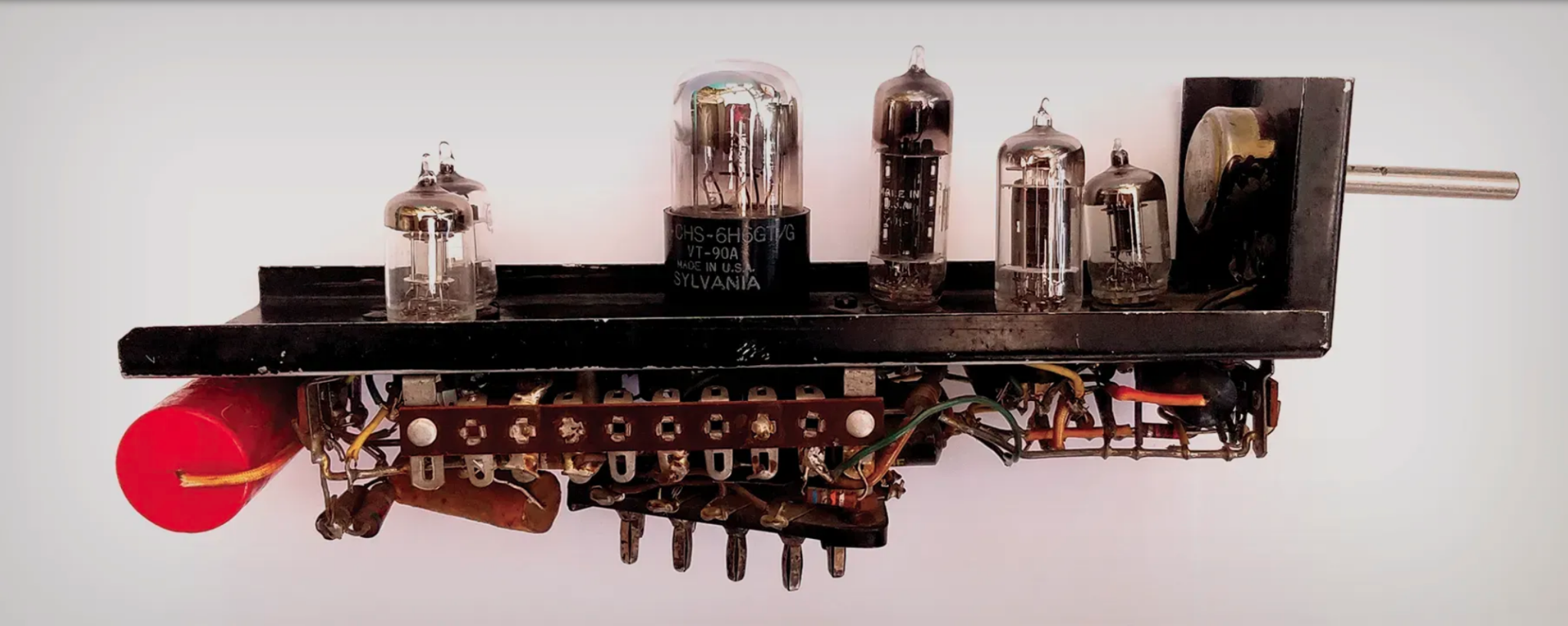}
	\caption{One node of 40 nodes constructing the stochastic neural-analog reinforcement calculator (SNARC) \cite{Akst2019-zi}}
	\label{fig:Fig5}
\end{figure}

Based on Alan Turing's "On Computable Numbers \cite{turing1937computable}," their model provided a way to abstractly describe brain
functions, and demonstrated that simple elements connected in a neural network can have enormous computational
power. The article received little attention until John von Neumann, Norbert Wiener, and others applied its concepts.
“McCulloch - Pitts'' neuron, was the first mathematical model of an artificial neural network. This model, inspired
by the basic physiology and function of the brain’s neurons, showed that essentially any computable function could
be modeled as a connected network of such neurons \cite{Russell2020-cx}. Based on this work, six years later, Donald Hebb proposed a
simple learning rule to tune the strength of the neuron connections \cite{hebb2005organization}. His learning method, namely “Hebbian
learning,” \cite{Song2000-bw} is considered as the inspiring model for neural networks learning. 
Building upon these works, one year later, in the summer of 1950, two Harvard undergrad students, Marvin Minsky and
Dean Edmonds built the first analog neural net machine called SNARC \cite{bernstein1981marvin}. SNARC stands for “stochastic
neural-analog reinforcement calculator” and was based on a network of 40 interconnected artificial hardware neurons
built using 3000 vacuum tubes and the remains of a B-24 bomber’s automatic pilot mechanism. SNARC was successfully
applied to find the way out from a maze (See figure~\ref{fig:Fig5}).

\begin{figure}[h]
    \centering
    \subfloat[\centering ]{{\includegraphics[width=8cm]{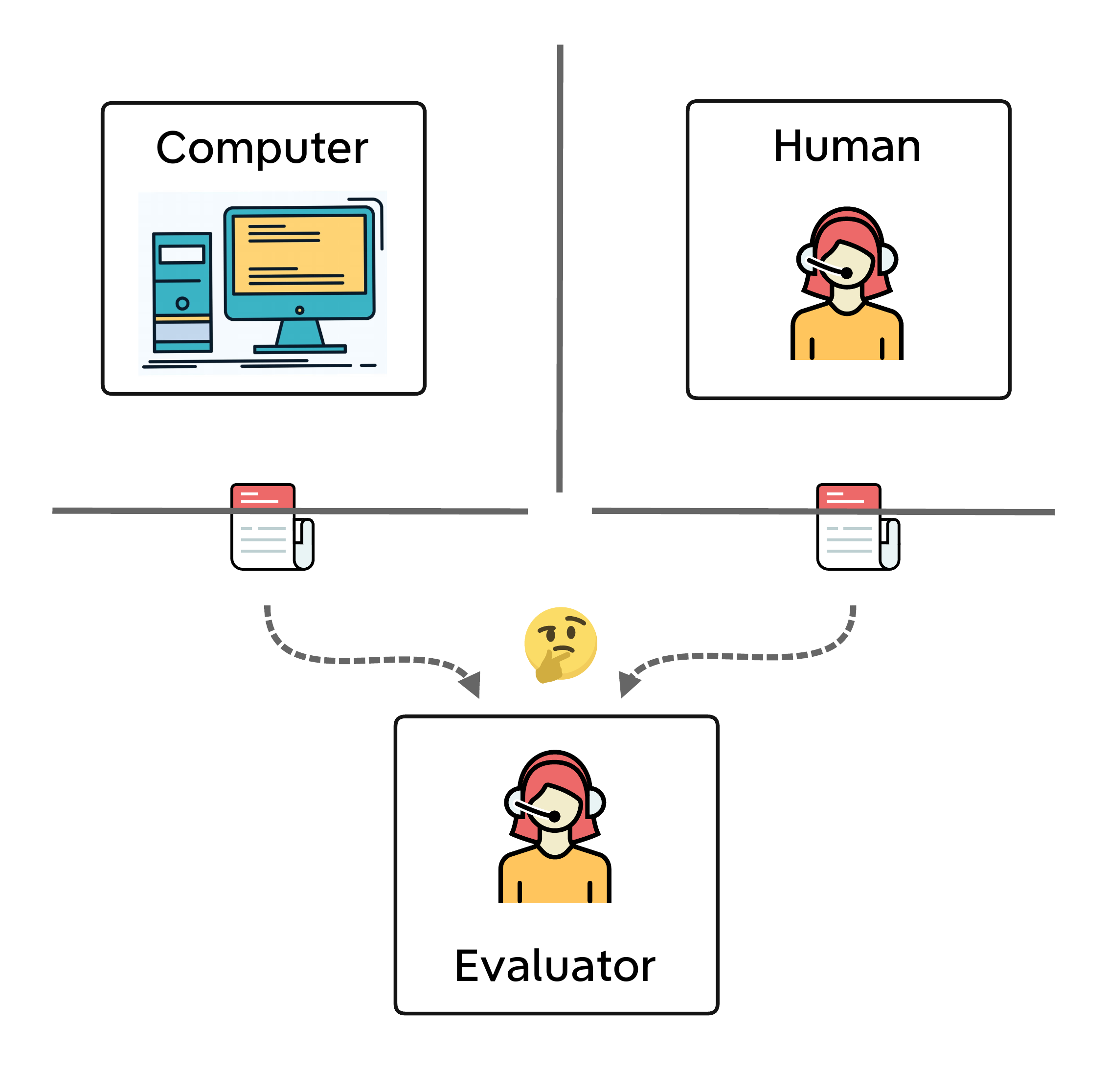} }}%
    \qquad
    \subfloat[\centering ]{{\includegraphics[width=6cm]{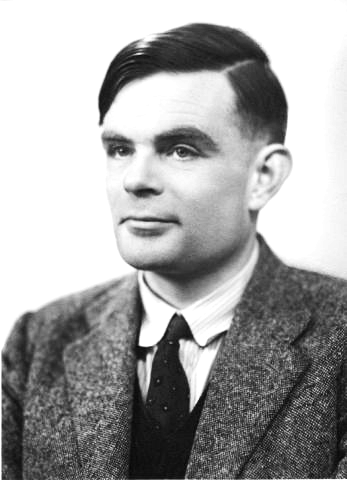} }}%
    \caption{(a): Alan Turing. (b): Schematic of the Turing Test.}%
    \label{fig:Fig6}%
\end{figure}

AI developed significantly from the studies of Alan Turing (figure~\ref{fig:Fig6}) during his short life, considered in all
respects as one of the fathers of AI. Although Turing owes much of his fame to the work he did at the Bletchley
Park center to decode German communications during World War II, his remarkable work toward the theory of
computation dates back to his article published when he was only 24 \cite{turing1937computable}. Turing demonstrated that his "universal
computing machine" could perform any imaginable mathematical computation if it could be represented as an
algorithm. John von Neumann stated that Turing's article laid the groundwork for the central concept of modern
computers. A few years later, in 1950, in his article entitled “Computing machinery and intelligence” \cite{Turing1950-xt}, Turing
raised the fundamental question of “Can a machine think?”. The imitation game or the Turing test evaluates the
ability of a machine to “think”. In this test, a human was asked to distinguish between a machine’s written answers
and those of a human (figure~\ref{fig:Fig6}). A machine is considered as being intelligent if the human interrogator could not
tell if the answer is given by a human or a machine \cite{noauthor_2020-di}. 

Before the term AI was coined, many works were pursued that were later recognized as AI, including two
checkers-playing games, developed almost at the same time by Arthur Samuel at IBM and Christopher Strachey at the
University of Manchester in 1952. 

\subsection{The First Summer of AI}
The term AI was coined around six years after Turing’s paper \cite{Turing1950-xt}, in the summer of 1956, when John McCarthy, Marvin
Minsky, Claude Shannon, and Nathaniel Rochester gathered common interest in automata theory, neural networks, and
cognitive science (2-month workshop at Dartmouth College). There, the term 'Artificial Intelligence' was coined by
McCarthy. McCarthy defined AI as "the science and engineering of making intelligent machines," emphasizing the
parallel growth between computers and AI. The conference is sometimes referred to as the "birthplace of AI" because
it coordinated and energized the field \cite{Russell2020-cx}, and this time is considered as the beginning of an era called “the first summer of AI”.

One of the consequent results of the Dartmouth Conference was the work of Newell and Simon. They presented a
mathematics-based system for proving symbolic logic theories, called the Logic Theorist (LT), along with a list
processing language for writing them called IPL (Information Processing Language) \cite{Crevier1994-px}. Soon after the conference,
their program was able to prove most of the theorems (38 out of 52 of them) in the second chapter of Whitehead \&
Russell’s “Principia Mathematica”. In fact, the program was able to give a solution for one of the theorems that
was shorter than the one in the text. Newell and Simon later released their General Problem Solver (GPS) which was
designed to mimic the problem-solving protocols of the human brain \cite{newell1959report}. General problem solver is counted as the first work in the
“reasoning humanly” framework of AI. 

\begin{figure}[h]
	\centering    
    \includegraphics[clip,width=1\linewidth]{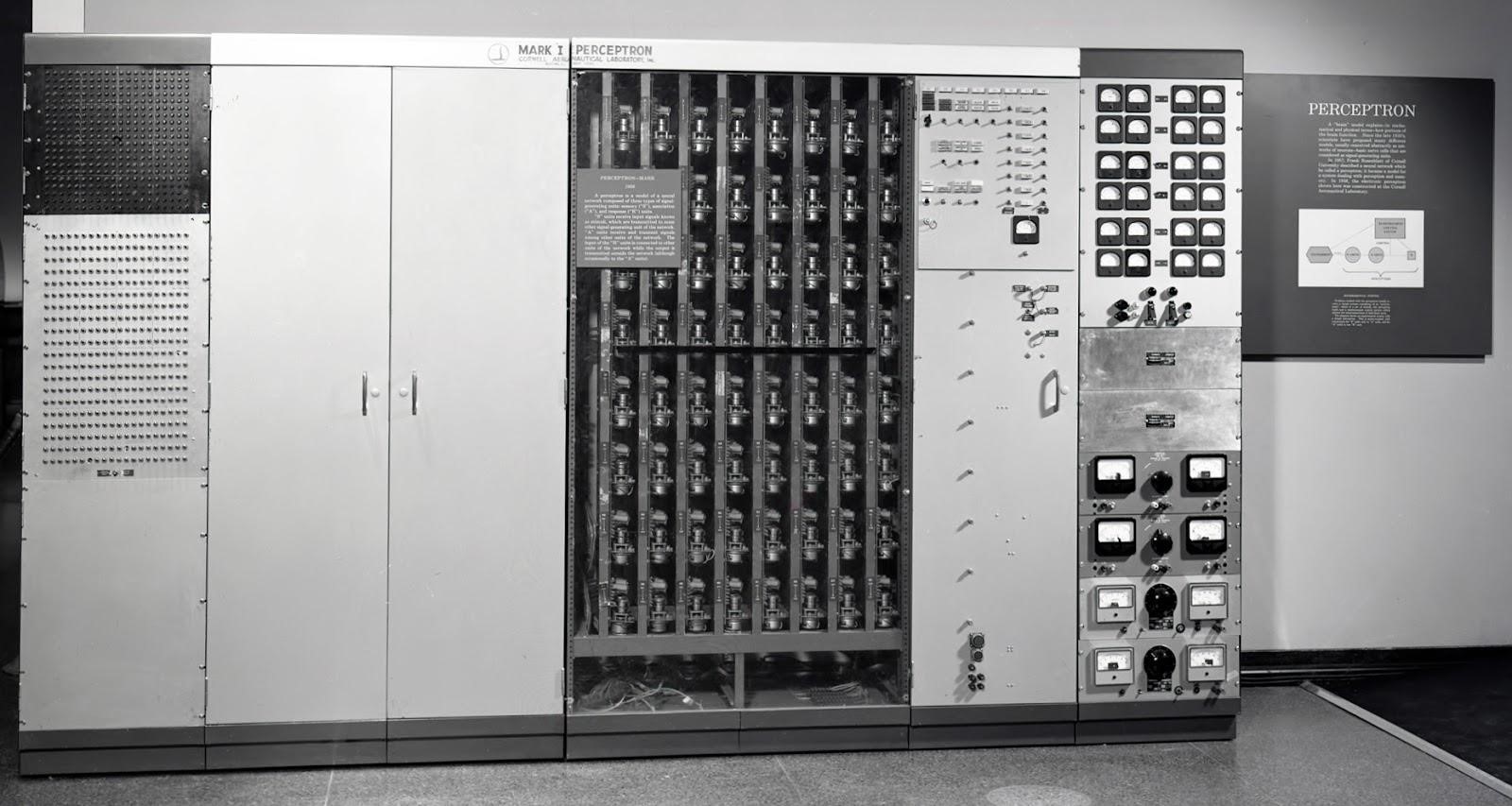}
	\caption{Mark I Perceptron at the Smithsonian museum \cite{noauthor_undated-iy}.}
	\label{fig:Fig7}
\end{figure}

Using reinforcement learning, Arthur Samuel’s 1956 checker player quickly learned to play at an intermediate level,
better than its own developer \cite{samuel1959some}. Reinforcement learning is a type of AI algorithm where an AI agent learns how
to interact with its surrounding environment to achieve its goal through a reward-based system. He demonstrated his
checker player program on television, making a great impression \cite{Bleakley2020-ot}. His work is considered to be the first
reinforcement learning-based AI program, and indeed the forefather of later systems such as TD-GAMMON in 1992, one
of the world’s best backgammon players \cite{Tesauro1995-gp}, and AlphaGo in 2016, which shocked the world by defeating the human
world champion of Go \cite{silver2016mastering}. A turning point in AI, and specifically in neural networks, occurred in 1957 when the
psychologist researcher Frank Rosenblatt (considered a father of deep learning \cite{Tappert2019-ac}) built the Mark I Perceptron at
Cornell \cite{Rosenblatt1957-bz}. He built an analog neural network with the ability to learn through trial and error. More precisely,
the perceptron was a single-layer neural network being able to classify the input data into two potential
categories. The neural network produces a prediction, say “left” or “right”, and if it is incorrect, it attempts to
get more accurate the following time. Accuracy increases with each iteration. A 5-ton IBM 704 computer the size of
a room was fed by a large stack of punch cards (figure ~\ref{fig:Fig7}). The computer learned to identify cards on the left and
cards on the right in 50 attempts. Mark I Perceptron is considered as one of the forefathers of modern neural
networks \cite{Melanie-Lefkowitz_perceptron}. 

In 1958 John McCarthy introduced the AI-specific programming language named LISP which became the prevailing AI
programming language for the next three decades. In his article entitled “Programs with Common Sense”, he proposed a
conceptual approach for AI systems based on knowledge representation and reasoning \cite{McCarthy1960-li}. LISP is the first
high-level AI programming language. In addition, 1958 is an important year because the first experimental work involving
evolutionary algorithms in AI \cite{De_Jong_undated-oj} was conducted by Freidberg toward automatic programming \cite{Friedberg1958-rn}. Nathaniel
Rochester and Herbert Gelernter of IBM developed a geometry-theorem-proving program in 1959. Their AI-based program
called “the geometry machine” was able to provide proofs for geometry theorems, which many math students had found quite
tricky \cite{Gelernter1960-ou}. Written in FORTRAN, their “geometry machine” program is regarded as one of the first AI programs that
could perform a task as well as a human. Another important event during the early 1960s is the emergence of the first
industrial robot. Named “Unimate”, the robotic arm was used on an assembly line in General Motors in 1961 for
welding and other metalworks \cite{Nof1999-uj}. 
In 1962, Widrow and Frank Rosenblatt revisited Hebb’s learning method. Widrow enhanced the Hebbian learning method
in his network called Adaline \cite{Widrow1960-yr} and Rosenblatt in his well-known Perceptrons \cite{Rosenblatt1957-bz}. Marvin Minsky in 1963
proposed a simplification approach for AI use cases \cite{Minsky1988-vm}. Marvin Minsky and Seymour Papert suggested that AI
studies concentrate on designing programs capable of intelligent behavior in smaller artificial environments. The
so-called blocks universe, which consists of colored blocks of different shapes and sizes arranged on a flat
surface, has been the subject of many studies. The framework called Microworld became a backbone for subsequent
works. Instances are James Slagle’s SAINT program for solving closed-form calculus integration problems in 1963
\cite{Slagle1963-wc}, Tom Evans’s ANALOGY in 1964 for solving an IQ test’s geometric problems \cite{Evans1964-zl}, and the STUDENT program, written in
LISP by Daniel Bobrow in 1967, for solving algebra problems \cite{bobrow1964natural}. ELIZA, the first chatbot in the history of AI was
developed by Joseph Weizenbaum at MIT in 1966. ELIZA was designed to serve as a virtual therapist to ask questions
and provide follow-ups in response to the patient \cite{Salecha2016-ql}. SHAKEY, the first omni-purpose mobile platform robot was
also developed at Stanford Research Institute in 1966 with reasoning about its surrounding environment \cite{noauthor_undated-xa}.

\subsection{The First Winter of AI}
The hype and high expectations caused by the media and the public from one side, and the false predictions and
exaggerations by the experts in the field about their outcome from the other side, led to major funding cuts in AI
research in the late 1960s. Governmental organizations like Defense Advanced Research Projects Agency (DARPA) had
already granted enormous funds for AI research projects during the 1960s. Two reports brought about major halts in
supporting the research: the US government report, namely ALPAC in 1966 \cite{Hutchins2003-pe}, and the Lighthill report of the
British government in 1973 \cite{Lighthill-Report-1972}. These reports mainly targeted the research pursued in AI, mostly the research works done
on artificial neural networks and came up with a grim prediction for the technology's prospects. As a result, both
the US and the UK governments started to decrease support for AI research at universities. DARPA, which had
previously funded various research projects in the 1960s, now needed specific timelines and concise explanations of
each proposal's deliverables. These events slowed the advancement of AI and ushered in the first AI winter, which
lasted until the 1980s. 

It is crucial to recognize the three key factors that caused this major halt in AI research for that era. First,
many early AI systems pursued the “thinking humanly” approach to solve the problems. In other words, instead of
taking a bottom-up approach starting from thoroughly analyzing the task, providing a possible solution, and turning
it into an implementable algorithm, they took the opposite direction, merely relying on replicating the way humans
perform the task. Second, there was a failure to recognize the complexity of many of the problems. Resulting from
the oversimplification of the AI frameworks proposed by Marvin Minsky, most early problem solving systems succeeded
mainly on toy (simplistic) problems, by combining simple steps to come up with a solution. However, many of the
real-world problems that AI was attempting to solve were in fact intractable. It was commonly assumed that “scaling
up” to bigger problems was merely a matter of faster hardware and greater memory capacity. However, developments in
computational complexity theory proved it wrong. The third factor was related to the negative conceptions about
neural networks and the limitations of their fundamental structures. In 1969 Minsky pointed out the limited
representational abilities of a perceptron, (to be exact, a single-layer perceptron cannot implement the classic
XOR logical function), and despite not being a general critique about neural networks, this also contributed to
global funding cuts in neural networks research. 

\subsection{Expert Systems; the Revival of AI, and the Second Summer}

Mainstream AI research efforts during the previous two decades were generally based on so-called “weak AI”, that
is, providing general solutions based on search algorithms in a space of all possible states built on basic
reasoning steps. Despite being general purpose, these approaches suffered from a lack of scalability to larger or
more complex domains. To address these drawbacks, in the early 1980s, researchers decided to take a more robust
approach utilizing domain-specific information for stronger reasoning but in narrower areas of expertise. The new
approach, so-called “expert systems”, originated at Carnegie Mellon University, and was quickly able to find its
way to corporations. DENDRAL \cite{feigenbaum1970-DENDRAL}, created at Stanford by Ed Feigenbaum, Bruce Buchanan, and Joshua Lederberg in the
late 1960s and early 1970s, and inferred molecular structure from mass spectrometry data, was an early success story.
DENDRAL was the first effective knowledge-intensive system, relying on a vast range of special-purpose laws, to
provide expertise rather than basic knowledge. 

In 1971 at Stanford University, Feigenbaum started the Heuristic Programming Project aimed at extending the area in
which expert systems could be applied. The MYCIN system was one of the successful consequent results of the new
wave, developed in the mid 1970s for the purpose of blood infection diagnosis by Edward Shortliffe under the
supervision of Bruce Buchanan and Stanley Cohen. MYCIN could perform identification of bacteria causing sepsis, and
recommend antibiotics dosage based on patient weight. It could perform diagnosis on par with the human experts in
the field, and significantly better than medical interns, benefiting from around 600 deduced rules in the form of a
knowledge base, from extensive interviews with the experts, by means of integrating uncertainty calculations \cite{Shortliffe1975-pk}.
Meanwhile, one of the most important moves toward deep convolutional neural networks (CNN) happened in 1980. The
“neocognitron”, the first convolutional neural networks (CNN) architecture, was proposed by Fukushima in 1980 \cite{Fukushima1980-gy}. Several learning algorithms
were suggested by Fukushima to train the parameters of a deep neocognitron so that it could learn internal
representations of input data. This work is in fact regarded as the origin of today’s deep convolutional neural networks.
R1, developed by McDermott in 1982, was the first successful commercial expert system used in the digital
equipment industry, for the configuration of new computer systems’ orders \cite{McDermott1985-hv}. In nearly 4 years, the firm added
\$40 million of revenue using R1. By 1988, most corporations in the United
States benefited from expert systems, either
by being a user of the system or doing research in the field \cite{Olsen1983-va}. The application of expert systems to real-world
problems resulted in the development of a wide range of representations and reasoning tools. The Prolog language
gained popularity in Europe and Japan, whereas the PLANNER language family thrived more in the United States. In Japan, the
government started a 10-year plan to keep up with the new wave by investing more than \$1.3 billion in
intelligent systems. The US government, by establishing the Microelectronics and Computer Technology Corporation in 1982, revived AI research in hardware, chip design, and software research. The same change happened
in the UK as well, resulting in reassignment of funds previously cut. All these events during the 80s led to a
period of “Summer” for AI. The AI industry thrived from billions of dollars invested in the field, and various
activities emerged from expert systems developer companies to domain-specific hardware, computer vision, and
robotic systems. Overall, the AI industry boomed from a few million dollars in 1980 to billions of dollars in the
late 1980s, including hundreds of companies building expert systems, vision systems, robots, and software and
hardware specialized for these purposes.

\subsection{The Second Winter of AI}
Despite all efforts and investments made during the early 1980s, many companies could not fulfill their ambitious promises. Hardware manufacturers declined to keep up with the requirements of specialized needs of the expert systems. Hence, the thriving industry of expert systems in the early 1980s declined tremendously and inevitably collapsed by the end of the 1990s and the AI industry faced another winter that lasted until the mid 1990s. This second period of so-called “winter” in the history of AI had been so harsh that AI researchers subsequently tended to avoid even the term “AI” by choosing other titles such as “informatics” or “analytics”. Despite the big shutdown of AI-based research works, the second winter was the time when the very well-known backpropagation algorithm was revisited by many research groups \cite{Rumelhart1986-vi} \cite{Ian2016-ek}. Backpropagation, which is a primary learning mechanism for artificial neural networks, was vastly used in learning problems during these years and eventually led to a new wave of interest in neural networks. The lesson learned during the periods of AI’s winter made researchers more conservative. As a result, during the late 1980s and the 1990s, the field of AI research witnessed a major conservative shift toward more established theories like statistics-based methods. Among these theories finding their way to the field were hidden Markov models (HMMs) \cite{Baum1966-bj}. Being strictly mathematical-based and resulting from extensive training on large real-world datasets, hidden Markov models became a trustable framework for AI research, especially in handwriting recognition and speech processing, helping them to make their way back to the industry. 
Another important outcome of this conservative shift in the field of AI was the development of public benchmark datasets and related competitions in its various subfields. Instances include the Letter Dataset \cite{Frey1991-zh}, Yale face database \cite{Georghiades1997-nx}, MNIST dataset \cite{lecun1998gradient}, Spambase Dataset \cite{dimitrakakis2002online}, ISOLET Dataset \cite{Fanty1990-bs}, TIMIT \cite{Zue1990-dk}, JARtool experiment Dataset \cite{Pettengill1991-ar}, Solar Flare Dataset \cite{Li2004-tu}, EEG Database \cite{Ingber1997-sj}, Breast Cancer Wisconsin (Diagnostic) Dataset \cite{Nick_Street1993-sh}, Lung Cancer Dataset \cite{Hong1991-za}, Liver Disorders Dataset \cite{Bagirov2003-xp}, Thyroid Disease Dataset \cite{Quinlan1987-ir}, Abalone Dataset \cite{Clark1996-js}, UCI Mushroom Dataset \cite{Iba1988-du} and other datasets that have been gathered during the 1990s. The availability of these public benchmarks became an important means for rigorous measurement of AI research advancements.

\subsection{Man versus Machine}
The gradual public interest in AI during the early 90s opened doors to other emerging or established fields such as
control theory, operational research, and statistics. Decision theory and probabilistic reasoning started being
adopted by AI researchers. Uncertainty was represented more effectively by introducing Bayesian networks to the
field \cite{Pearl2009-mm}. Rich Sutton in 1998 revisited reinforcement learning after around thirty years by adopting Markov
decision processes \cite{sutton2018reinforcement}. This step led to a growth in applying reinforcement learning on various problems, such as
planning research, robotics, and process control. The vast amount of available data in different areas on the one
hand, and the influences of statistical methods such as machine learning and optimization on AI research methods,
on the other, resulted in significant readoption of AI in subfields including multiagent models, natural language
processing, robotics, and computer vision. As such, new hopes for AI shaped again in the early 1990s. Eventually, in
1997, AI-equipped machines showed off their power against “Man” to the public \cite{Weber1997-iv}. Chess-playing AI software
developed in IBM, called “Deep Blue”, eventually won over the great maestro chess world champion, Garry Kasparov.
Broadcasted live, Deep Blue captured the public’s imagination once again toward AI systems of the future. The news
was so breathtaking that IBM’s share values rose up to all-time highs \cite{Higgins2017-ta}. 

\subsection{Information Age: Enter Big Data}
Massive advances in microchip manufacturing technologies in the late 1990’s led to emerging powerful computers,
concurrent to the growth of the global Internet that generated massive amounts of data. This information included enormous
unprocessed text, video, voice, and images, along with semiprocessed data such as geographical tracking, social
media-related data, and electronic medical records, ushering in the era of big data \cite{Morris2018-tx}. In the computer vision
area in 2009, the ImageNet dataset was created gathering millions of labeled images, significantly contributing to
the field \cite{Gershgorn2017-bf}. There was a new beginning of wide interests in AI from the industry. Notable steps were taken in
2011 when IBM’s Watson defeated human champions in the highly popular TV quiz show Jeopardy \cite{Gabbatt2011-nm},
significantly boosting public impression of the state-of-the-art in AI, and with the introduction of Apple’s Siri
intelligent assistant. 

\subsection{Return of Neural Networks}
\begin{figure}[h]
	\centering    
    \includegraphics[clip,width=1\linewidth]{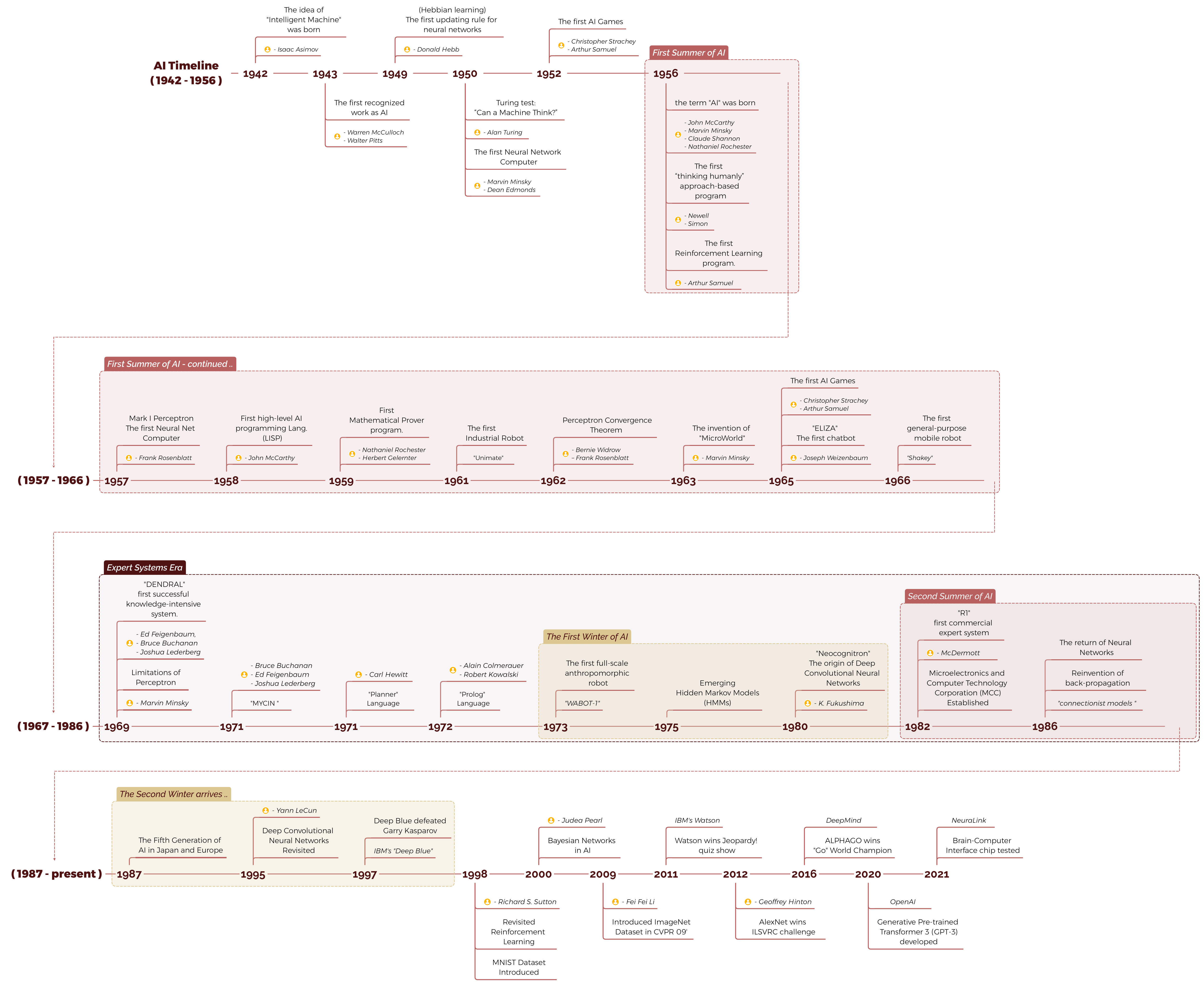}
	\caption{The timeline of developments in AI}
	\label{fig:Fig8}
\end{figure}

In 1989 Yann LeCun revisited convolutional neural networks, and using gradient descend in their training mechanism,
demonstrated the ability to perform well in computer vision problems, specifically in handwritten digit recognition
\cite{lecun1989backpropagation}. Yet, it was in 2012 that these networks came to the forefront. A deep convolutional neural network developed
in Geoffrey Hinton’s research group at the University of Toronto surpassed the ImageNet Large Scale Visual
Recognition Challenge (ILSVRC) competitors by significantly enhancing all ImageNet classification benchmarks \cite{Imagenet-ILSVRC2012}. Before
the use of deep neural networks, all best-performing methods were mostly so-called classical machine
learning methods using hand-crafted features. By 2011 the computing power of graphics processing units had grown enough to help the researchers train deep networks with higher dimensions both in terms of width and
depth in a shorter time. Since the earlier implementation of Convolutional Neural Networks on graphics processing units in 2006 \cite{Chellapilla2006-ev}
which resulted in four times faster performance compared with central processing units, Schmidhuber’s team at IDSIA could achieve a 60 times
faster performance on graphics processing units in 2011 \cite{Ciresan2011-dt}. Meanwhile, the availability of huge amounts of labeled data such as
millions of labeled images in the ImageNet dataset helped researchers to overcome the problem of overfitting.
Eventually, in 2012 Hinton’s team proposed a deep convolutional neural network architecture, named AlexNet (after
the team’s leading author Alex Krizhevsky), which was able to train more layers of neurons. Using many
mechanisms and techniques such as rectified linear unit activation functions and the dropout technique, AlexNet
could achieve higher discriminative power in an end-to-end fashion, that is, to feed the network with merely the
pure images of the dataset \cite{Krizhevsky2012-pm}. This event is regarded as the birth of the third boom in AI. Since then, deep
learning-based methods have continued to achieve outstanding feats, including outperforming or performing on par
with human experts in certain tasks. Instances include AI-related fields such as computer vision, natural language
processing, medical image diagnosis \cite{Liu2019-na}, and natural language translation. The progress of deep neural networks
gained public attention in 2016 at the time when Deep Mind’s AlphaGo beat the world champion of Go \cite{silver2016mastering}. AI became
again the target of the media, public, governments, industries, scholars, and investor’s interests. Deep learning
methods have nowadays entirely dominated AI-related research, creating entirely new lines of research and
industries. In 2018, Yoshua Bengio, Geoffrey Hinton, and Yann LeCun won the Turing award for their pioneering
efforts in deep learning. Figure~\ref{fig:Fig8} summarizes the timeline of AI from the time it was born up to now.

\section{Where Do We Stand Now?}

The previous section presented a brief story of AI’s journey, with all its ups and downs over the decades. This
journey has not been easy, with multiple waves and seasons. Specifically, AI has faced two main breakdowns (so-called
winters) and three main breakthroughs (so-called AI summers or booms). Thanks to the convergence of parallel
processing, higher memory capacity, and more massive data collection (e.g, big data), AI has enjoyed a steady upward
climb since the early 2020s. With all these pieces in place, much better algorithms have been developed, assisting this
steady progression.
Computers are becoming faster. Computing power has continued to double nearly every two years (Moore’s law).
Advancements in technology occur 10 times faster, that is, what used to take years, now may happen in the course of
weeks or even days \cite{Manson2016-hp}.
On a global scale, AI is becoming an attractive target for investors, producing billions of dollars of profit per
annum. From 2010 to 2020, global investment in AI-based startup companies has steadily grown from \$1.3 billion to more than
\$40 billion, with an average annual growth rate of nearly 50\%, whereas in 2020 only, corporate investment in AI is reported
to be nearly \$70 billion globally \cite{Raymond_Perrault2019-ol}.
In the academic sector, from 2000 up until 2020, the number of peer-reviewed AI articles per year has grown roughly 12
times worldwide. AI conferences have witnessed similar significant increases in terms of the number of attendants. In
2020, NeurIPS accepted 22,000 attendees, more than 40\% growth over 2018, and 10-fold more compared with 2012. 
Concurrently, AI has become the most popular specialization among computer science PhD students in North America,
nearly three times the next rival (theory and algorithms) \cite{aiindex2021}. In 2019, more than 22\% of PhD candidates in
computer science majored in AI and machine learning. 
With the introduction of machine learning, the environment of the health care and biology sectors has changed
dramatically. AlphaFold, developed by DeepMind, used deep learning to make a major advance in the decades-long biology
problem of protein folding. Scientists use machine learning algorithms to learn representations of chemical molecules
to plan more efficient chemical synthesis. Machine learning-based approaches were used by PostEra, an AI startup, to speed up
coronavirus disease 2019-related drug development during the pandemic \cite{aiindex2021}.
This progress suggests that we are in the midst of the next hype cycle. And this new hype is focused on applications with
life-or-death implications, such as autonomous vehicles, medical applications, and so on, making it critical that AI
algorithms be trustworthy.
\section{The Future of AI}

Numerous AI-related startups have been founded in recent years, with both companies and governments investing
heavily in the sector. If another AI winter occurs, many will lose their jobs, and many startups will be forced to
close, as has occurred in the past. According to McKinsey \& Company, the economic gap between an approaching winter
period and continued prosperity by 2025 would be in the tens of billions of dollars \cite{Chui2018-sp}.
A recurrent pattern in previous AI winters has been the promises that sparked initial optimism yet turned out to be
exaggerated. During both AI winters, budget cuts had a major effect on AI research. The Lighthill report resulted in
funding cuts in the UK during the first AI winter, as well as cuts in Europe and the United States.
DARPA support was cut, resulting in the second AI winter. Significant attention needs to be paid to technical
challenges and limitations. Let us recall what was faced by the perceptron in the 1960s in being noted as unable to
solve the so-called XOR problem, or limitations faced by expert systems in the 1980s. AI has appeared particularly
vulnerable to overestimations coupled to technical limitations. Overall, the hype and fear that comes with reaching
human-level intelligence have quickly contributed to exaggerations and public coverage that is not common in other
innovative tech sectors. To avoid a next winter of AI, a number of important considerations may need to be made:

\begin{enumerate}[label=\roman*.,leftmargin=*,align=parleft]
\item It is extremely important to be aware of philosophical arguments about the utter sublimeness of what it means to
be human, and to not make exaggerated claims about ascension of AI systems to being human (see very illustrating
documentary \cite{Ruspoli2018-fx}). In addition, these philosophical arguments (e.g. by Hubert Dreyfus based on the philosophy of
Martin Heidegger), had they been more extensively and interactively considered, could have likely contributed to
further success by AI in early years (e.g. earlier attention to ‘connectionist’ approaches to AI).

\item Neglects regarding above point, as well as exciting early successes, contribute to exaggerated claims that AI
will solve any important problem soon. This is what happened that contributed to the first winter of AI, where AI
researchers made overconfident and overoptimistic predictions about upcoming successes, given the early promising
performances of AI on simpler examples \cite{Taylor_undated-en}. A lack of appreciation for the computational complexity theory was
another reason for AI scientists to believe that scaling up of simple solutions to larger tasks is just a matter of
using faster hardware and larger memories. 

\item According to a recently released report \cite{Vincent2019-hg}, 40\% of startups established in Europe that claim to use AI in
their provided services do not actually do so, largely because the definition of AI is
ambiguous for the majority of the public and the media. Therefore, given recent excitement around AI and the
resulting hype and investment growth in the field, some businesses try to benefit from this ambiguity by misusing
terms such as AI, machine learning, and deep learning. Thus, it is crucial to define these terms more clearly with
respect to other related concepts and define what they have in common and what they do not. 

\item There is significantly troubled trends in scientific methodology and dissemination by AI researchers,
contributing to the hype and confusion: these trends include failure to distinguish between explanation and
speculation, failure to identify real sources of performance gains, confusing/misleading use of math, and misuse of
language \cite{lipton2019research}. According to a recent study \cite{Roberts2021-ez} reviewing a spectrum of machine learning approaches for detecting
and prognosticating coronavirus disease 2019 from standard-of-care chest radiographs and computed tomography images,
none of the more than 400 studies were found suitable for clinical application! The studies suffered from one or multiple issues
including use of poor-quality data, poor application of machine learning methodology, poor reproducibility and
biases in study design.

\item AI, in its essence, is vague and covers a broad scope. As observed by Andrew Moore \cite{Irving-Wladawsky-Berger}, “Artificial intelligence
is the science and engineering of making computers behave in ways that, until recently, we thought only human
intelligence is able to perform”. The critical point in this definition lies in the phrase "until recently" which
points to the moving target of AI through time. In other words, ideas and methods are being referred to as AI as
long as they have not been completely discovered. Once they are figured out, they may be no longer associated with
AI and receive their own tag. This phenomenon is known as the AI effect \cite{noauthor_undated-zi} which contributes to the fast decline
in public excitement about ground breaking achievements in AI. 

\item An important challenge with AI technologies, more specifically deep learning-based AI, is its so-called
black box, opaque nature of decision making. That is, when a deep learning algorithm makes a decision, the process
of its inference or the logic behind it may not be representable. Although in some tasks such as playing board
games, for example, “Go”, where the objective is merely winning the game, this concern does not reveal itself, in critical
tasks; for example in health care, where a decision impacts humans lives, this issue can lead to trustworthiness challenges
(other examples of critical tasks include transportation and mobility systems, and social or financial systems.)

\item One very specific concern is issue of bias. As an example, soon after Google introduced bidirectional encoder representations from transformers (BERT), one of the most sophisticated AI technologies in language models,
scientists learned an essential flaw in that system. BERT and its peers (GPT-2, GPT-3, t5, etc) were more likely to
equate males with computer programming and, in general, failed to give females adequate respect. BERT, which is now
being used in critical services such as Google's Internet search engine, is one of a number of AI systems (so-called
“transformers”) that learn from massive amounts of digitized data, such as old books, Wikipedia pages, tweets,
forums, and news stories. The main problem with BERT or generative pretrained
transformers (GPT-3), BERT’s rival
introduced by OpenAI, and similar universal language models is that they are too complex even for their own
designers (GPT-3's full version has 175 billion parameters \cite{Brown2020-us}); in fact, scientists are still learning how these
models work. One certain fact about these systems is that they pick up biases as they learn through human-generated
data. Because these mechanisms can be used in a variety of sensitive contexts to make critical and life-changing
choices, it is critical to ensure that the decisions do not represent biased attitudes against particular
communities or cultures. Thus, builders of AI systems have a duty to steer the design and use of AI in ways that
serve society.
\end{enumerate}

Overall, the future of AI seems to be promising. AI eventually will drive our automobiles, aid physicians in making more
precise diagnoses, assist judges in making more consistent judgments, help employers in hiring more qualified
applicants, and much more. We are aware, however, that these AI systems may be fragile and unjust. By adding
graffiti to a stop sign, the classifier may believe it is no longer a stop sign. By adding subtle noise or signal to
a benign skin lesion image, the classifier may be fooled into believing it is malignant (eg, adversarial attacks).
Risk management instruments used in US courts have been shown to be racially discriminatory. Corporate recruitment
tools have been shown to be sexist.
Toward trustworthy AI, organizations around the world are coming together to establish consistent standards for
evaluating responsible implementation of AI systems and to encourage international support for AI technologies that
benefit humanity and the environment. Among these tries is the European Commission’s report on Ethics guidelines
for trustworthy AI \cite{Floridi2019-nz} and DARPA’s XAI (eXplainable AI) roadmap \cite{Gunning2019-kx}. 

\begin{figure}[h]
	\centering    
    \includegraphics[clip,width=0.55\linewidth]{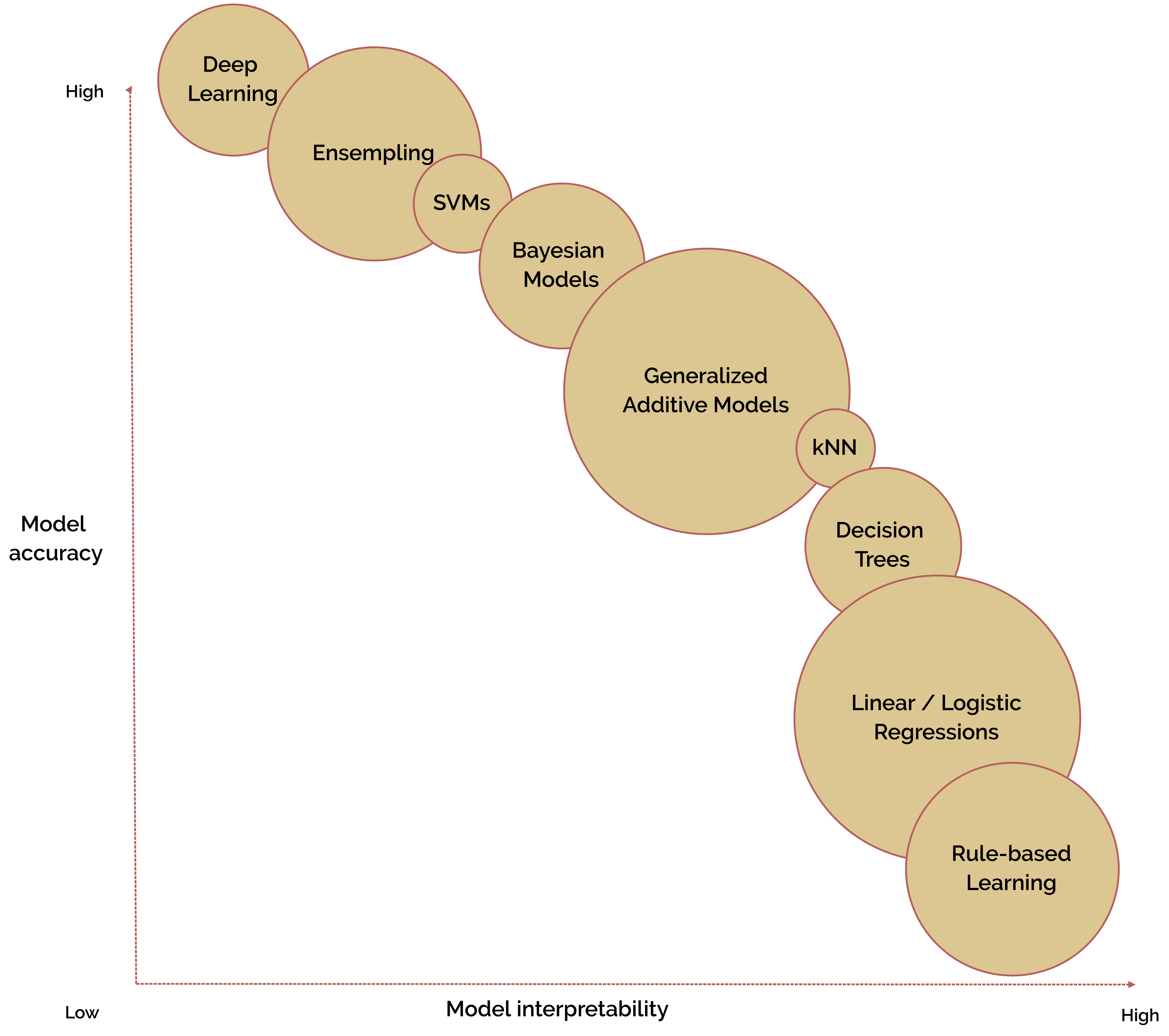}
	\caption{Trade-off between interpretability and performance for AI models. (Adapted from Arrieta, Alejandro Barredo, et al. “Explainable Artificial Intelligence (XAI): Concepts, taxonomies, opportunities and challenges toward responsible AI.” Information Fusion 58 (2020): 82-115. With permission. (Figure 12 in original).)}
	\label{fig:Fig9}
\end{figure}

\begin{figure}[H]
	\centering
    \includegraphics[clip,width=1\linewidth]{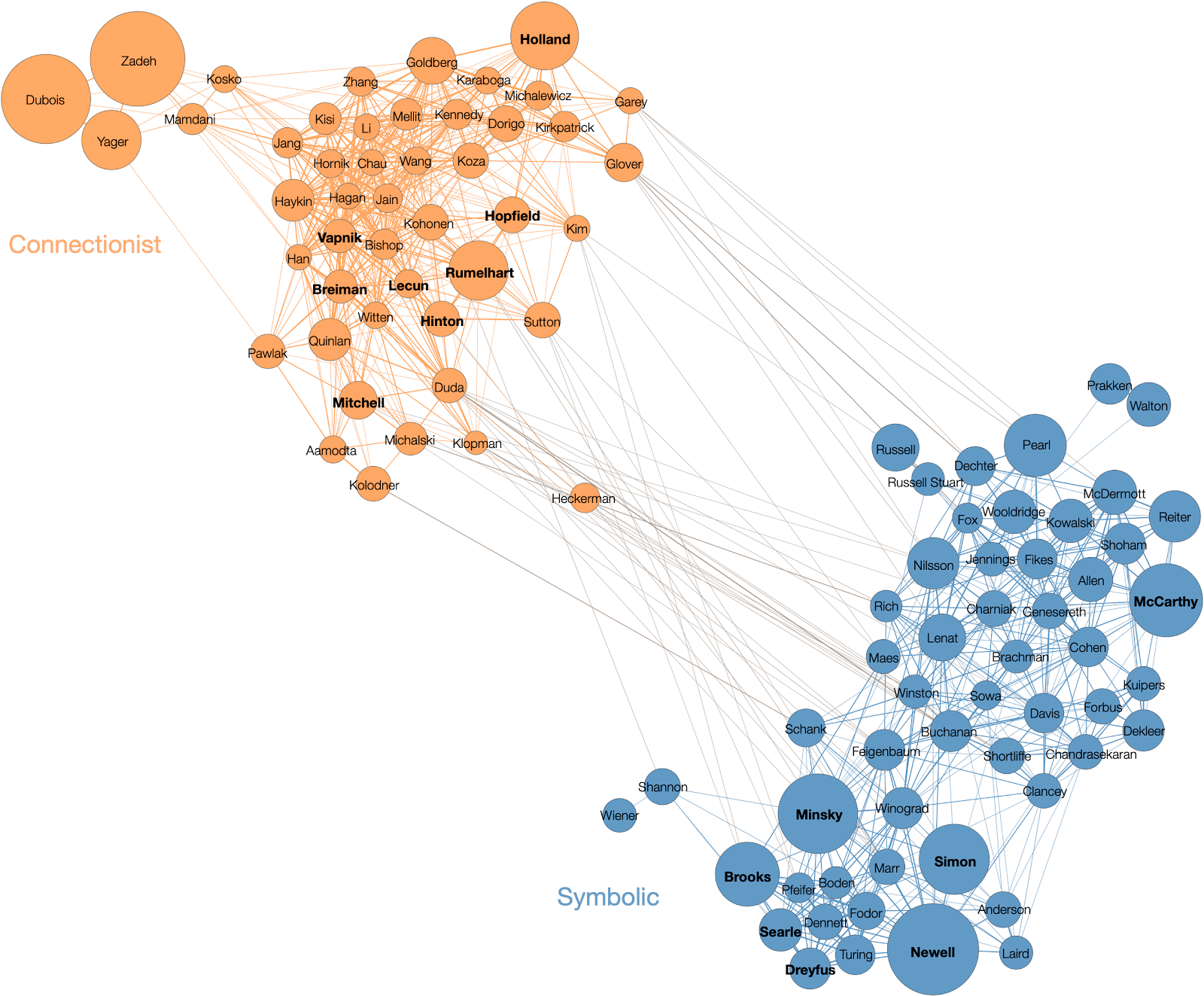}
	\caption{Co-citation network of the 100 most cited authors with “Artificial Intelligence” in the title. Figure illustrates the names of some important authors, clearly distributed by the community. At the heart of the “connectionists,” some core figures in deep learning appear. On the “symbolic side,” some core figures are laid out in a way that represents their proximities and divergences, surrounded by primary contributors to the construction of cognitive modeling, expert systems, and even those critical of symbolic AI (Dreyfus, Searle, Brooks). This figure is not comprehensive and misses some key contributors; and is intended to demonstrate existing dichotomies in connectionist and symbolic frameworks.(From Cardon, Dominique, Jean-Philippe Cointet, and Antoine Mazières. “Neurons spike back. The invention of inductive machines and the artificial intelligence controversy”, Réseaux, vol. 211, no. 5, 2018, pp. 173-220. Available at: \url{https://neurovenge.antonomase.fr/}. Accessed May 18 2021; with permission.)}
	\label{fig:Fig10}
\end{figure}

According to Arrieta and colleagues, \cite{Barredo_Arrieta2020-lu}, a trade-off between interpretability of AI models and their accuracy (performance) can be observed, given fair comparison conditions (figure ~\ref{fig:Fig9}). Simpler AI approaches, such as linear regression and decision trees, are self-explanatory (interpretable) because the classification decision border may be depicted in a few dimensions using model parameters. However, for tasks such as categorization of medical images in health care, these may lack the necessary complexity, yet to acquire the trust of physicians, regulators, and patients, a medical diagnostic system needs to be visible, intelligible, and explainable; it should be able to explain the logic of making a certain decision to stakeholders engaged in the process. Newer rules, such as the European General Data Protection Regulation, are making the use of black box models more difficult in different industries because retraceability of judgments is increasingly required. An AI system designed to assist professionals should be explainable and allow the human expert to retrace their steps and use their judgment. Some academics point out that humans are not always competent or willing to explain their choices. However, explainability is a fundamental enabler for AI deployment in the real world because it ensures that technology is used in a safe, ethical, fair, and trustworthy manner. Breaking AI misconceptions by demonstrating what a model primarily looked at while making a judgment can help end-users to trust the technology (eg, via use of heat maps/activation maps). For non-deep learning users, such as most medical professionals, it is even more vital to show such domain-specific attributes used in the decision. 
For further enhanced AI, a way forward seems to be the convergence of symbolic and connectionist methods, which would incorporate the former's higher interpretability with the latter's significant recent success (more on these next). For instance, the use of hybrid distributional models, which combine sparse graph-based representations with dense vector representations and connect those to lexical tools and knowledge bases, appears to be promising toward explainable AI in the medical domain \cite{Gunning2017-in}. However, the main obstacle toward this solution is the historical division between these two paradigms. 
The deep neural network approach is not novel. Today, it is fulfilling the promise stated at the beginning of cybernetics by benefitting from developments in computer processing and the existence of massive datasets. These techniques, however, have not always been deemed to constitute AI. Machine-learning approaches based on neural networks (connectionist AI) have been historically scorned and ostracized by the symbolic school of thinking. The rise of AI, which was clearly distinct from early cybernetics, amplified the friction between these two approaches. The cocitation network of the top-cited authors in articles mentioning AI demonstrates the drift between researchers who have used the symbolic or connectionist paradigms (See figure~\ref{fig:Fig10}). 

Despite the obvious separation that existed among the intellectuals from these two schools, a third subfield of AI has been emerging, namely neuro-symbolic AI, which focuses on combining the neural and symbolic traditions in AI for additional benefit \cite{Sarker2021-mu}. The promise of NeSy AI is largely based on the aim of achieving a best-of-both-worlds scenario in which the complementary strengths of neural and symbolic techniques can be advantageously merged. On the neural side, desirable strengths include trainability from raw data and robustness against errors in the underlying data, whereas on the symbolic side, one would like to retain these systems' inherent high explainability and provable correctness, as well as the ease with which they can be designed and function using deep human expert knowledge. In terms of functional features, using symbolic approaches in conjunction with machine learning –particularly deep learning, which is currently the subject of the majority of research– one would hope to outperform systems that rely entirely on deep learning on issues such as out-of-vocabulary handling, generalizable training from small datasets, error recovery, and, in general, explainability \cite{Sarker2021-mu}.

\section{Summary}

Rapid developments in AI are changing different aspects of human life. Advances both in computational power and AI algorithm design have enabled AI methods to outperform humans in an increasing number of tasks. AI has experienced decades of praise and criticism; its path has never been smooth. With the two winters that the field has experienced, after two waves of great growth and high expectations, as well as the costs that the community of researchers, corporations, start-ups, and governments have paid, it is critical for us to recognize that the current wave of high hopes and high expectations should not be taken for granted.

\section*{Acknowledgment}

This work was in part supported by the Canadian Institutes of Health Research (CIHR) Project Grant PJT-162216. The authors also wish to acknowledge valuable feedback from Ian Janzen of BC Cancer Research Institute.

\footnotesize
\RaggedRight
\bibliographystyle{unsrt}
\bibliography{biblio}
\section*{Glossary}

\footnotesize
\begin{longtable}{p{.20\textwidth} p{.75\textwidth}}
\textbf{Term:}                                          &\textbf{Definition:}   \\
 
\hline
 
  & \\
  
 Artificial Intelligence (AI) 
 & The study and development of computer systems capable of imitating intelligent human behavior. \\
 
  & \\
  
 Machine learning (ML) 
 & A subset of AI in which computers learn how to do tasks through the use of massive amounts of data rather than being programmed. \\
 
  & \\
  
 Artificial neural networks (ANN) 
 & A type of computer system that is supposed to operate in a manner comparable to the human brain and nervous system. \\
 
  & \\
  
 Convolutional neural networks (CNN) 
 & A type of feed-forward neural network that consists of a large number of convolutional layers stacked on top of one another. It is primarily used for computer vision tasks. \\
 
  & \\
  
 Deep learning (DL)
 & A type of computer system that is supposed to operate in a manner comparable to the human brain and nervous system. \\
 
  & \\
  
 Reinforcement learning (RL) 
 & A subfield of machine learning that studies how intelligent entities should behave in order to optimize the concept of cumulative reward. \\
 
  & \\
   
 Hidden Markov model (HMM)
 & A type of Markov model in which the represented system is considered to be a Markov process.\\
 
  & \\
 
 General problem solver (GPS)
 & A computer software designed in 1959 to act as a universal problem solver. \\
 
 & \\
 
 Bayesian Networks
 & A probabilistic graphical model that utilizes a directed acyclic graph (DAG) to describe a set of variables and their conditional dependencies. \\
 
 & \\
 
 Perceptron 
 & A supervised learning approach for binary classifiers in machine learning. \\
 
 & \\
 
 Backpropagation
& Most popular method for training feedforward neural networks. \\

& \\
 
 Gradients Vanishing
 & When backpropagation is used to train artificial neural networks, the gradient can become vanishingly small in some situations, thereby preventing the weight from changing its value. \\
 
 & \\
 
 Natural language processing (NLP) 
 & Series of methods for processing of natural languages, such as translating. \\
 
 & \\
 
 Computer vision 
 & Computer vision is an interdisciplinary scientific topic that examines how computers can extract information from digital images or video streams at a high degree of understanding. \\
 
 & \\
 
 Speech Recognition 
 & Refers to the technology that enables computers to comprehend spoken language. \\
 
 & \\
 
 cognitive science 
 & The scientific study of the mind and its processes on an interdisciplinary level. \\
 
 & \\
 
 logical notation 
 & Refers to a collection of symbols that are frequently employed to convey logical representations. \\
 
 & \\
 
 Turing test 
 & A test of a machine's ability to demonstrate intelligent behavior that is comparable to, or indistinguishable from, human behavior. \\
 
 & \\
 
 CVPR 
 & The Conference on Computer Vision and Pattern Recognition (CVPR), is an annual conference on computer vision and pattern recognition. \\
 
 & \\
 
 AAAI 
 & The Association for the Advancement of AI, an international scientific society devoted to promoting research in, and responsible use of AI. \\
 
 & \\
 
 AI INDEX Report 
 & The AI Index is a public-facing annual study on the state of AI in all relevant fields. \\
 
 & \\
 
 XOR function 
 & A logical "exclusive OR" operator that returns TRUE if one of the logical propositions is true and FALSE if both statements are true. It also returns FALSE if neither of the statements is true. \\
 
 & \\
 
 strong AI 
 & A machine described as being capable of applying intelligence to any challenge. \\
 
 & \\
 
 weak AI
 & Used in contrast to "strong AI," which is described as a machine capable of applying intelligence to any problem, rather than just one specific problem. \\
 
 & \\
 
 expert systems
 & A computer program that simulates the decision-making abilities of a human expert. It is supposed to handle difficult problems through the use of bodies of knowledge, which are mostly represented as if–then rules. \\
 
 & \\
 
 Microelectronics and Computer Technology Corporation (MCC) 
 & One of the largest computer industry research and development consortia in the United States. \\

 & \\

 MNIST dataset 
 & The Modified National Institute of Standards and Technology database (MNIST) is a massive collection of handwritten digits that is frequently used to train image processing systems. \\
 
 & \\
 
 ImageNet 
 & A big visual database that was created for the purpose of doing research on visual object recognition software. \\
 
 & \\
 
 Decision Theory 
 & The study of an agent's decisions is called decision theory. It is closely related to the discipline of game theory and is researched by economists, statisticians, data scientists, psychologists, biologists, social scientists, philosophers, and computer scientists. \\
 
 & \\
 
 Markov Decision Processes 
 & A discrete-time stochastic control process that provides a mathematical framework for modeling decision-making in situations where outcomes are partly random and partly under control. \\
 
 & \\
 
 Big Data 
 & A branch of study that focuses on methods for analyzing, extracting information from, or otherwise dealing with data volumes that are too vast or complicated for typical data-processing application software to handle. \\
 
 & \\
 
 Data science 
 & An interdisciplinary field that uses scientific methods, procedures, algorithms, and systems to mine organized and unstructured data for information and insights. \\
 
 & \\
 
 ILSVRC 
 & The ImageNet Large Scale Visual Recognition Challenge is an annual software competition in which participants compete to classify and recognize objects and scenes properly. \\
  
 & \\
 
 GPU 
 & A graphics processing unit (GPU) is a specialized electronic circuit that is capable of swiftly manipulating and altering memory in order to expedite the production of images in a frame buffer for output to a display device. \\
  
 & \\
 
 ReLU (Rectified Linear Unit) 
 & The rectifier, or ReLU (Rectified Linear Unit), is an activation function whose positive portion is specified as the argument's positive component. \\
  
 & \\
 
 Deep Mind 
 & Deep Mind is an artificial intelligence company and research laboratory of Alphabet Inc. based in the United Kingdom. It was created in September 2010 and was bought by Google in 2014. \\
  
 & \\
 
 Moore’s law 
 & Moore's law is a historical observation and projection of a pattern in which the number of transistors in a dense integrated circuit (IC) doubles approximately every two years.
 \\
  
 & \\
 
 NeurIPS 
 & NeurIPS, The Conference and Workshop on Neural Information Processing Systems, is an annual conference on machine learning and computational neuroscience held in December. \\
  
 & \\
 
 WiML workshop 
 & The annual WiML Workshop is a technical event where women can share their machine learning research. \\
  
 & \\
 
 AI4ALL 
 & AI4ALL is a not-for-profit organization dedicated to advancing diversity and inclusion in the field of artificial intelligence. \\
  
 & \\
 
 AlphaFold 
 & AlphaFold is a machine learning algorithm built by Google's DeepMind that predicts the structure of proteins. \\
  
 & \\
 
 BERT 
 & Bidirectional Encoder Representations from Transformers (BERT) is a Google-developed machine learning methodology for pre-training in natural language processing (NLP).
 \\
  
 & \\
 
 GPT-3 
 & Generative Pre-trained Transformer 3 (GPT-3) is a deep learning-based autoregressive language model that generates human-like writing. \\
  
 & \\
 
 XAI 
 & XAI is a type of AI in which the solution's outcomes are understandable by humans. It contrasts with the concept of the "black box" in machine learning, in which even the designers of the AI are unable to explain why it made a certain decision. \\
 
 & \\
  
   
\end{longtable}

\end{document}